\documentclass[a4paper,fleqn]{cas-dc}

\usepackage[numbers,sort&compress]{natbib}
\usepackage{subcaption}
\usepackage{graphicx}
\usepackage{booktabs}
\usepackage{multirow}
\usepackage{makecell}
\usepackage{array}
\usepackage{algorithm}
\usepackage{algpseudocode}
\usepackage{placeins}

\begin{document}
\let\WriteBookmarks\relax
\RenewDocumentCommand \printorcid { } { }
\renewcommand{\topfraction}{0.95}
\renewcommand{\bottomfraction}{0.90}
\renewcommand{\textfraction}{0.05}
\renewcommand{\floatpagefraction}{0.80}
\renewcommand{\dbltopfraction}{0.95}
\renewcommand{\dblfloatpagefraction}{0.80}
\setcounter{topnumber}{5}
\setcounter{bottomnumber}{5}
\setcounter{totalnumber}{8}

\shorttitle{Adaptive Time-step Training for Spike-Based NeRF}
\shortauthors{Lin et al.}

\title[mode=title]{Adaptive Time-step Training for Enhancing Spike-Based Neural Radiance Fields}

\author[1]{Ranxi Lin}
\credit{Conceptualization, Methodology, Software, Validation, Formal analysis, Investigation, Visualization, Writing -- original draft}
\ead{linrx2024@shanghaitech.edu.cn}

\author[2]{Canming Yao}
\credit{Validation, Formal analysis, Data curation, Writing -- review \& editing}
\ead{yaocanming@gdiist.cn}

\author[1]{Jiayi Li}
\credit{Software, Validation, Investigation, Data curation}
\ead{lijy22023@shanghaitech.edu.cn}

\author[1]{Weihang Liu}
\credit{Software, Validation, Investigation, Data curation}
\ead{liuwh2023@shanghaitech.edu.cn}

\author[1]{Xin Lou}
\credit{Supervision, Project administration, Writing -- review \& editing}
\ead{louxin@shanghaitech.edu.cn}

\author[1]{Pingqiang Zhou}
\credit{Supervision, Project administration, Writing -- review \& editing}
\cormark[1]
\ead{zhoupq@shanghaitech.edu.cn}

\affiliation[1]{organization={School of Information Science and Technology, ShanghaiTech University},
            city={Shanghai},
            country={China}}

\affiliation[2]{organization={Guangdong Institute of Intelligence Science and Technology},
            city={Zhuhai},
            country={China}}

\cortext[1]{Corresponding author}


\begin{abstract}
Spiking Neural Networks (SNNs) provide an energy-efficient computing paradigm for neural rendering, but existing spike-based Neural Radiance Field (NeRF) models usually use a fixed inference time step for all scenes. This fixed temporal budget is inefficient because NeRF follows a scene-specific training paradigm, and different scenes require different temporal capacities to preserve rendering quality. This paper proposes Pretraining-based Adaptive Time-step Adjustment (PATA), a scene-wise adaptive time-step training framework for spike-based NeRF. PATA parameterizes the target inference time step as a trainable variable and optimizes it through a two-stage training process. A hybrid input mode strengthens early time-step outputs, while full-step soft supervision, smoothed rendering loss, and temporal-budget loss jointly maintain rendering fidelity and reduce temporal computation. The learned target time step is shared by all ray samples within a scene, preserving the parallel rendering structure of NeRF. Experiments on INGP-NeRF and TensoRF backbones across Synthetic-NeRF, Mip-NeRF 360, and LLFF show that PATA consistently reduces inference cost while maintaining competitive rendering quality. PATA reduces the estimated inference energy by up to 57.57\% on INGP-NeRF and 68.90\% on TensoRF, demonstrating its effectiveness across different neural rendering representations.
\end{abstract}

\begin{highlights}
\item PATA learns scene-wise temporal budgets for spike-based NeRF rendering.
\item A two-stage objective optimizes rendering quality and inference time steps.
\item PATA generalizes across hash-grid and tensor-factorized NeRF backbones.
\item PATA reduces estimated inference energy by up to 68.90\% on evaluated benchmarks.
\end{highlights}


\maketitle
\section{INTRODUCTION}

Neural Radiance Fields (NeRF) \cite{mildenhall2021nerf} has been widely used in various 3D reconstruction and rendering tasks with superior performance. By encoding scene geometry and appearance implicitly through a neural network, NeRF maps 3D coordinates and viewing directions to volumetric density and view-dependent radiance. This representation enables high-fidelity image synthesis from arbitrary viewpoints. However, training NeRF requires dense sampling of points along rays cast from multiple camera poses, leading to extensive computational demands. The predominant use of Multi-Layer Perceptrons (MLPs) \cite{zhu2024vanilla} further exacerbates this issue, as the network must process a massive number of multiply-accumulate (MAC) operations per sample, resulting in substantial computational load and energy consumption \cite{barron2021mip, yao2023spikingnerf, liu2024content}.

While prior work has explored pruning \cite{zhang2025asfc, xie2023hollownerf} and quantization \cite{hasssan2025quant, zhou2025gradient} to reduce computational demand during inference, these methods still rely primarily on multiply-accumulate (MAC)-based computation. In contrast, Spiking Neural Networks (SNNs) offer a paradigm shift by leveraging event-driven computation and binary spike communication. Unlike conventional Artificial Neural Networks (ANNs), SNNs replace MAC operations with sparse accumulations (AC), and their spiking neurons activate only when membrane potentials exceed a threshold, naturally enforcing activation sparsity. This biologically inspired architecture not only enhances computation and energy efficiency but also enables native processing of spatiotemporal data \cite{tavanaei2019deep, rathi2023exploring}. Building on these advantages, recent studies have successfully integrated SNNs with NeRF for 3D rendering tasks, demonstrating both high rendering quality and significantly reduced energy consumption \cite{liao2024spiking, li2025spiking, wang2025mixed, gu2024sharpening, guo2024spike, yao2023spikingnerf}.

\begin{figure*}[!t]
  \centering
  \begin{subfigure}[b]{0.25\textwidth}
    \includegraphics[width=\textwidth]{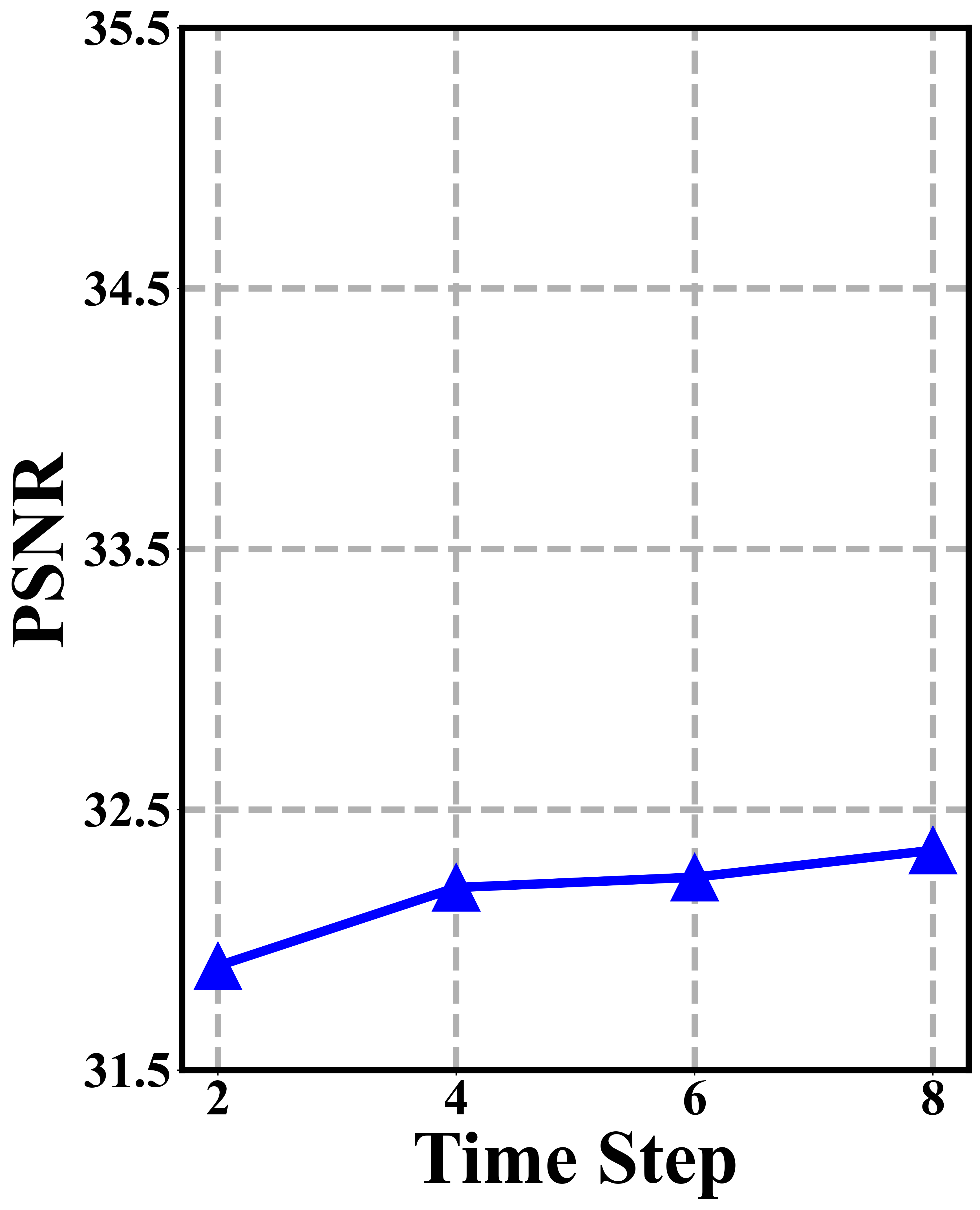}
    \caption{Ficus}
  \end{subfigure}
  \hfill
  \begin{subfigure}[b]{0.25\textwidth}
    \includegraphics[width=\textwidth]{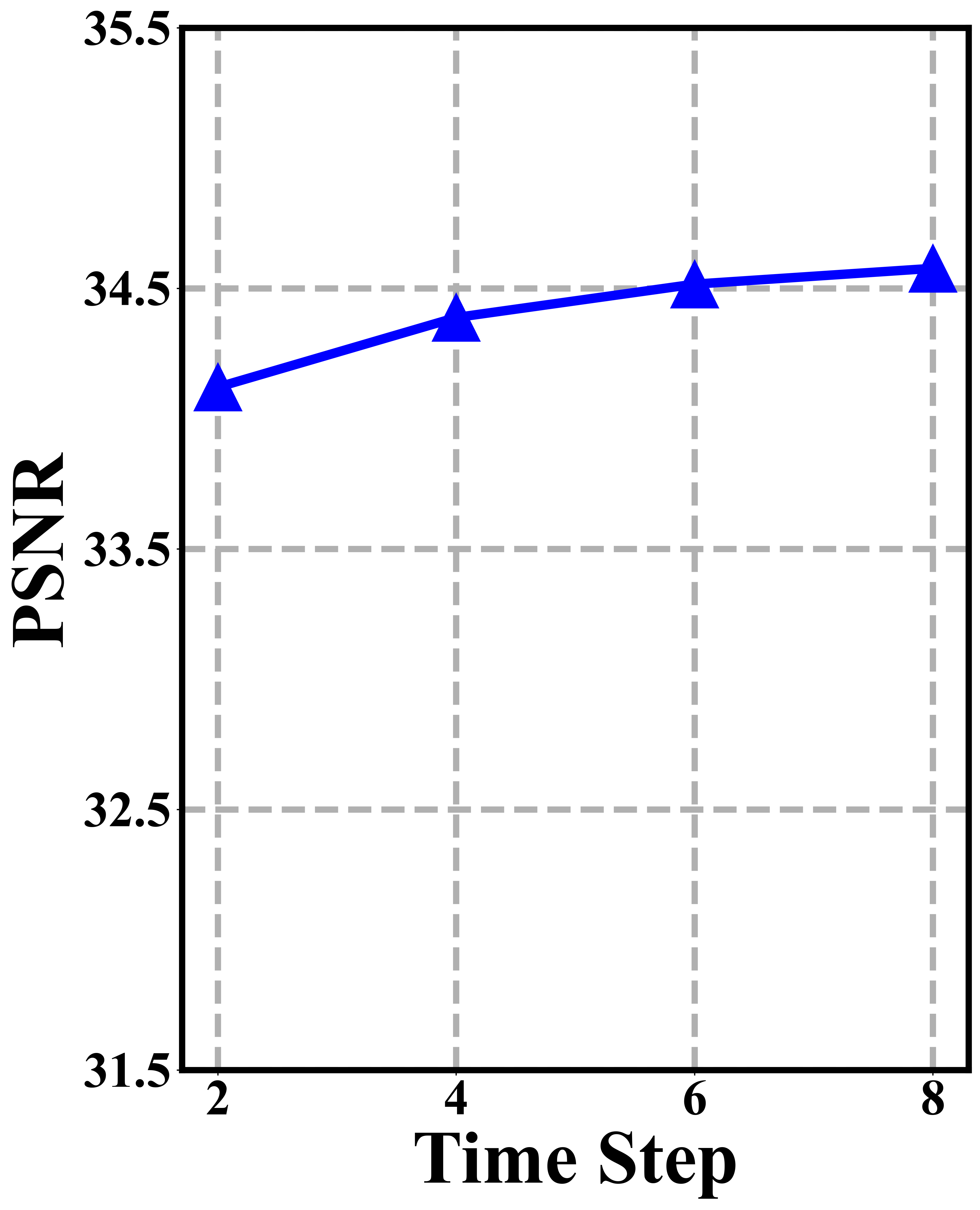}
    \caption{Chair}
  \end{subfigure}
  \hfill
  \begin{subfigure}[b]{0.25\textwidth}
    \includegraphics[width=\textwidth]{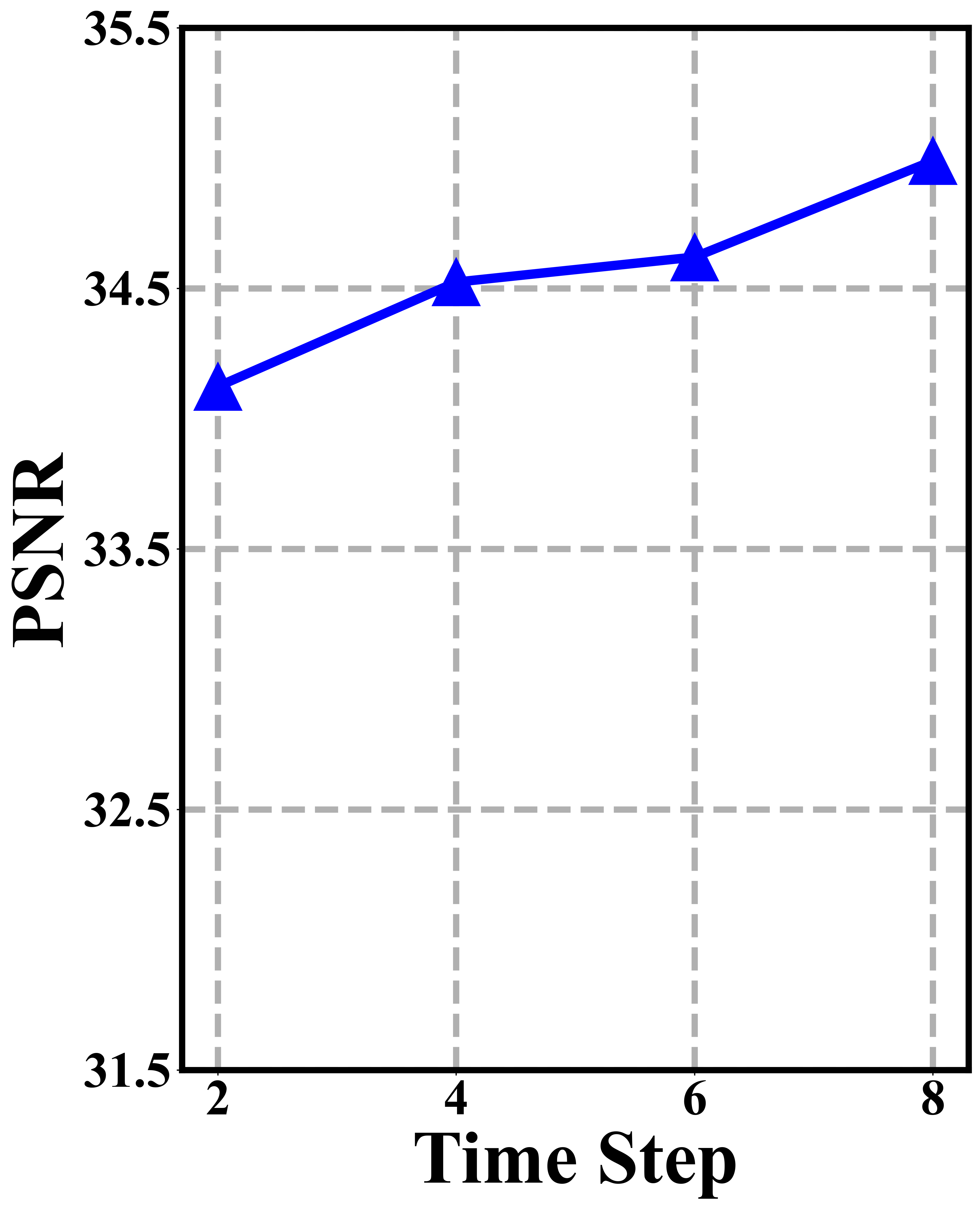}
    \caption{Mic}
  \end{subfigure}
  \caption{PSNR increases with the inference time step across representative scenes, while different scenes reach comparable quality at different temporal budgets.}
  \label{fig:psnr}
\end{figure*}

However, spike-based NeRF models introduce a temporal dimension that is absent from conventional ANN-based NeRF. Prior SNN studies have shown that increasing the time step $T$ improves temporal information capacity~\cite{datta2022can, kim2023exploring}, while NeRF itself follows a scene-specific training paradigm in which the required model capacity varies with geometry and texture complexity~\cite{liu2024content,takikawa2022variable,girish2023shacira,yang2022recursive}. These two properties make a fixed inference time step inefficient for spike-based NeRF. As shown in Figure~\ref{fig:psnr}, representative scenes reach comparable rendering quality at different temporal budgets.


This temporal-budget problem differs from adaptive time-step learning in conventional SNN tasks. Existing dynamic inference strategies usually make decisions at the sample level or layer level, such as confidence-based early stopping~\cite{li2023unleashing, li2023input, zhong2024towards} or layer-wise temporal allocation~\cite{li2023seenn}. These designs are not well matched to neural rendering, where a scene is rendered by processing massive ray samples in a dense and highly parallel manner. If different samples or layers terminate at different time steps, the resulting irregular execution can weaken the parallel rendering pattern. This motivates a scene-wise temporal-budget strategy: each scene learns one target time step, and all ray samples within that scene share the same temporal budget during inference.


A practical temporal-budget strategy should also avoid being tied to a single NeRF backbone. 
Modern neural rendering systems use diverse scene representations, including hash-grid encodings~\cite{muller2022instant} and tensor-factorized fields~\cite{tensorf}. 
Therefore, an adaptive time-step method for spike-based NeRF should answer two questions: how to learn a scene-specific temporal budget without manual search, and whether the learned strategy remains effective when the underlying neural rendering representation changes.

To address these questions, we propose Pretraining-based Adaptive Time-step Adjustment (PATA), a dynamic time-step training strategy for spike-based NeRF models. 
PATA parameterizes the target inference time step and optimizes it jointly with rendering quality through a two-stage training procedure. 
To strengthen early time-step predictions, we introduce a hybrid input mode that uses different input attenuation strategies across time steps. 
We further use the full-step output as soft supervision for the target-step prediction and apply a smoothing strategy to stabilize training when the target time step changes. 
During inference, the trained model renders each scene using its learned target time step, reducing temporal computation while preserving visual fidelity.




Our main contributions are summarized as follows:
\begin{itemize}
    \item[$\bullet$] \textbf{Scene-wise Temporal-Budget Learning}: We formulate time-step selection in spike-based NeRF as a learnable scene-wise temporal budget, avoiding manual time-step search while preserving the parallel rendering pattern required by dense ray sampling.

    \item[$\bullet$] \textbf{Stable Adaptive Time-Step Training}: We introduce PATA, a two-stage training strategy that combines hybrid input dynamics, target-step parameterization, full-step soft supervision, and smoothing regularization to jointly optimize rendering quality and inference time steps.

    \item[$\bullet$] \textbf{Backbone-General Validation}: We validate PATA on both INGP-NeRF and TensoRF backbones, showing that the proposed temporal-budget learning strategy transfers across hash-grid and tensor-factorized neural rendering representations while maintaining a controllable quality-efficiency trade-off.
\end{itemize}

\section{RELATED WORK}
\subsection{Neural Radiance Fields}

Since the introduction of Neural Radiance Fields (NeRF), significant research efforts have been devoted to enhancing its capabilities from multiple perspectives \cite{barron2021mip,shin2023binary, reiser2024binary, muller2022instant, chen2024far, jiang2024g}. 
The original NeRF framework revolutionized novel view synthesis by employing a neural network to implicitly encode scene properties, mapping 3D coordinates and viewing directions to volumetric density and view-dependent radiance. 
This approach achieves photorealistic rendering through differentiable volume rendering.
Several notable improvements have been made to the original NeRF architecture. Mip-NeRF \cite{barron2021mip} addresses aliasing artifacts by introducing multiscale scene representation and conical frustum sampling, significantly improving rendering quality. 
TensoRF~\cite{tensorf} adopts a tensor-factorized representation of radiance fields, decomposing scene information into low-rank tensor components to reduce memory and computation while preserving rendering quality. 
For computational efficiency, Instant-NGP \cite{muller2022instant} proposes a hash-based feature grid representation combined with two compact MLPs, dramatically accelerating rendering while maintaining quality. In the domain of model compression, 
BiRF \cite{shin2023binary} implements a binary-encoded feature grid through binarization-aware training, achieving substantial memory savings without reducing computational complexity during inference.

\subsection{Direct Training of SNNs}

SNNs typically process information through multiple time steps, requiring efficient information flow across both spatial (layers) and temporal (time steps) dimensions during training and inference. The non-differentiable nature of spiking operations poses a fundamental challenge for gradient-based optimization. To address this, Wu et al.\cite{wu2018spatio} proposed Spatio-Temporal Backpropagation (STBP) with surrogate gradients, which has become a standard training approach for SNNs \cite{neftci2019surrogate}. Recent advances have focused on two main directions: (1) enhancing neuronal dynamics through modified computational processes \cite{yao2022glif, yin2021accurate, liao2024spiking, lian2024lif, zhang2025lif}, often at the cost of increased computational overhead; (2) developing specialized loss functions that leverage SNN characteristics to improve training efficiency \cite{deng2022temporal, yu2025efficient, yan2022backpropagation, liu2024teas, zuo2024self}, which can optimize network sparsity and architecture \cite{yan2022backpropagation, zuo2024self} without additional computational complexity.

\subsection{Spike-based Neural Rendering}

Recent advances have demonstrated successful integration of SNNs with NeRF \cite{liao2024spiking, li2025spiking, wang2025mixed, gu2024sharpening, guo2024spike, yao2023spikingnerf}. 
Liao et al. \cite{liao2024spiking} developed a hybrid ANN-SNN architecture with novel spiking neuron dynamics to enable continuous geometric representation. 
For geometric reconstruction tasks, Gu et al. \cite{gu2024sharpening} introduced an innovative polling strategy and designed specialized surrogate gradients. 
Yao et al. \cite{yao2023spikingnerf} addressed novel view synthesis by mapping ray sampling points to SNN time steps and proposing a TCP scheme to handle variable sampling densities across rays. 
SAH-NeRF~\cite{chang2026sah} explores an SNN-ANN hybrid framework for improving NeRF-based novel-view synthesis. 
Wang et al. \cite{wang2025mixed} further enhanced the framework through a grouped spiking neuron mechanism for improved cross-region communication.
Compared to these approaches, our solution offers distinct advantages: (1) it maintains the original neuron computational model without introducing additional neuronal operations, (2) it learns a scene-specific temporal budget during training instead of relying on a fixed or manually selected inference time step, and (3) it preserves the input/output dimensionality of the underlying NeRF backbone, making it compatible with existing spike-based neural rendering architectures.

\subsection{Dynamic Time Step}
Dynamic time-step adaptation has been widely studied to boost the efficiency of SNNs~\cite{li2023unleashing, li2023input, zhong2024towards, li2023seenn, datta2025dynamic, XINGTING2026108350}. Representative methods include confidence-based early stopping~\cite{li2023unleashing, li2023input, zhong2024towards}, reinforcement-learning-based per-sample adjustment~\cite{li2023seenn}, and mask-based layer-wise optimization for Spiking Transformers~\cite{datta2025dynamic}. These methods mainly target classification or sequence tasks, where the adaptive unit is typically an input sample, token, or layer.

Neural rendering has a different computational structure. A NeRF model is trained separately for each scene and renders images by processing massive ray samples in parallel. Under this workload, sample-wise or layer-wise temporal decisions may introduce irregular execution and weaken the parallel rendering pattern. Therefore, we adopt the scene as the adaptation unit: each scene learns a target time step that reflects its rendering complexity, while all ray samples in the scene share a consistent temporal budget during inference.

\begin{figure*}[!t]
    \centering
    \includegraphics[width=0.97\textwidth]{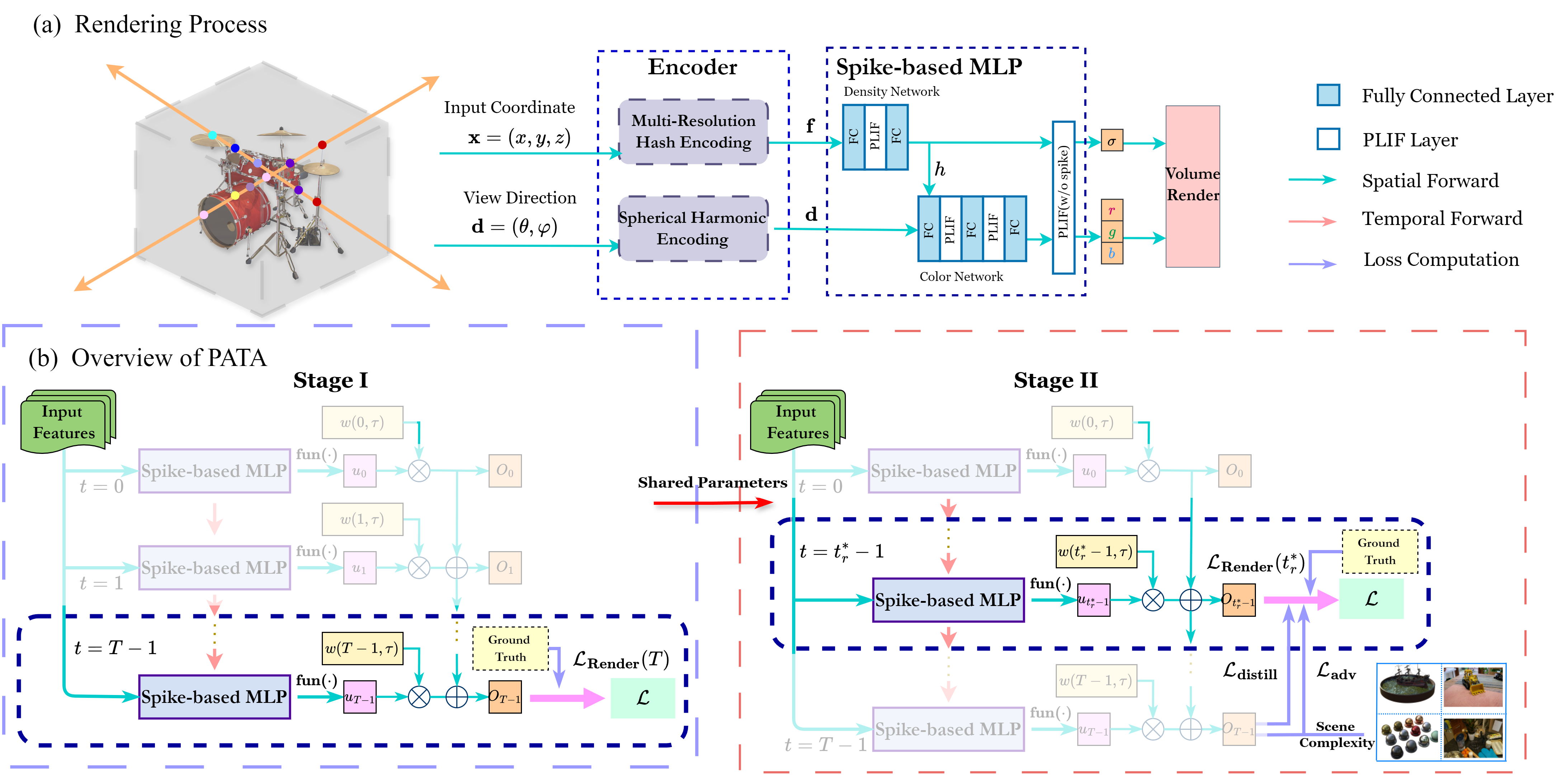}
    \caption{Overview of the proposed PATA framework with INGP-NeRF as the illustrative backbone. 
            (a) Spike-based neural rendering replaces the nonlinear activations in the density and color MLPs with spiking neurons, producing time-indexed density $\sigma$ 
            and color $\mathbf{c}$ for volume rendering. (b) PATA first trains a full-step renderer at the maximum time step $T$, then optimizes a learnable target time 
            step $t^*$ using smoothed rendering, distillation, and temporal-budget losses. PATA operates on temporal density/color outputs before volume rendering, 
            allowing the same mechanism to be applied to other NeRF backbones.}
    \label{fig:net_and_pata}
\end{figure*}

\section{PRELIMINARIES}

\subsection{Grid-based Neural Radiance Field} \label{ngp}
The INGP-NeRF model divides space into multiple grids of different resolutions, assigning indices $\mathbf{x}$ and feature vectors $\Phi_{\theta}$ to the corners of each voxel within these grids, which are then stored in a hash table. For the coordinates of input sample points, linear interpolation is performed based on their relative positions within different hierarchical grids to obtain the feature vectors of the sample points, which are then fed into the density network. For the viewing direction, it is encoded using the Spherical Harmonics (SH) encoder before being fed to the color network:
\begin{gather}
    \mathbf{f} = \mathrm{interp}(\mathbf{x}, \Phi_{\theta}) \\
    h, \sigma = MLP_{density}(\mathbf{f}), \;
    \mathbf{c} = MLP_{color}(\mathbf{d}, h)
\end{gather}
INGP-NeRF employs multiscale occupancy grids to expedite the sampling process. During training, it dynamically labels whether each grid cell is nonempty. At the time of sampling, only nonempty grid cells are sampled, markedly enhancing sampling efficiency. After the aforementioned process has taken place, for the sampled points lying on the same ray $ \mathbf{r}(t) = \mathbf{o} + t \mathbf{d}$, the RGB value of the ray is calculated using the volume rendering method proposed in \cite{mildenhall2021nerf}:
\begin{gather}
    \alpha_i = 1 - exp(-\sigma_i \delta_i), \; 
    T_i = \prod _{j = 1}^{i-1}(1 - \alpha_j)   \label{pointsweight} \\
\hat C(\mathbf{r} ) = \sum_{i=1}^{N}T_i\alpha_i\mathbf{c}_i \label{pointscolor}
\end{gather}
where $T_i$ represents the transmittance and $\delta_i = t_{i+1} - t_i$ denotes the distance between adjacent sampled points. 
We use INGP-NeRF as the main backbone to describe the spike-based rendering pipeline in Figure~\ref{fig:net_and_pata}. 
This choice provides a clear instantiation of the feature encoding, density prediction, color prediction, and volume rendering process. 
It should be noted that PATA does not rely on the hash-grid representation itself. Instead, it operates on the temporal outputs of the density and color prediction modules before volume rendering. 
Therefore, other NeRF backbones with the same density-color rendering interface can also be combined with PATA, as further evaluated with TensoRF in the experiments.

\subsection{Spiking Neuron Model}
The most widely used neuron model in SNNs is currently the Leaky Integrate-and-Fire (LIF) model \cite{zhang2024tc}. The membrane potential update equation is as follows \cite{fang2023spikingjelly}:
\begin{equation}
v^l(t) = 
\begin{cases}
    (1-\frac{1}{\tau})v^l(t-1) +\frac{1}{\tau}i(t), & if \;  decay \;  input;\\
    (1-\frac{1}{\tau})v^l(t-1) +i(t), & if \; not \;  decay \; input; 
\end{cases}\label{inputmode}
\end{equation}
where $i(t) = W^ls^{l-1}(t)$, with $s^{l-1}(t)$ representing the output spikes of neurons in the preceding layer at time t, $v^l(t)$ denotes the membrane potential of neurons in the l-th layer at time t, $\tau$ is the decay factor, which governs the decay amplitude of the membrane potential at each time step. If the decaying input mode is employed, a fraction of the input will undergo decay before accumulating; otherwise, the input will be directly added to the membrane potential. When the membrane potential exceeds the threshold, the LIF neuron will emit a spike and reset. We opt for the soft reset with the calculation carried out as follows:
\begin{equation}
v^l(t) = v^l(t) - s^l(t)\theta^l
\end{equation}
where $s^l(t) = H(v^l(t) - \theta^l)$, with $H(\cdot)$ denoting the Heaviside function. Due to the non-differentiability of the Heaviside function, the most widely used solution to enable parameter updating in SNNs using backpropagation algorithms is the surrogate gradient. In our experiments, the surrogate gradient used is as follows:
\begin{equation}
\begin{aligned}
\frac{\partial H(x)}{\partial x}
&\approx
\alpha \sigma(\alpha x)\left(1-\sigma(\alpha x)\right),\\
\sigma(\alpha x)
&=\frac{1}{1+\exp(-\alpha x)}
\end{aligned}
\label{eq:surrogate_gradient}
\end{equation}
where $\alpha$ is a predefined hyperparameter and defaults to 4.

\section{THE PROPOSED METHOD}

In this section, we introduce the proposed Pretraining-based Adaptive Time-step Adjustment (PATA) strategy, which effectively reduces the inference time steps and temporal computation required for rendering. The complete framework, including both the radiance field reconstruction and training procedure, is depicted in Figure~\ref{fig:net_and_pata}.

\subsection{Spike-based Neural Rendering}

Figure~\ref{fig:net_and_pata}(a) illustrates the spike-based neural rendering pipeline used by PATA. Given the scene features produced by an underlying NeRF backbone, we replace the nonlinear activations in the density and color prediction networks with Parametric Leaky Integrate-and-Fire (PLIF) neurons~\cite{Fang_2021_ICCV}. The PLIF neurons share a trainable decay factor $\tau$ within each spiking layer. To make rendering available at different time steps, the output layers accumulate membrane potentials without spike emission, producing time-indexed estimates of density $\sigma$ and color $\mathbf{c}$. These estimates are then passed to the differentiable volume rendering module. This design preserves the input-output interface of the underlying neural rendering backbone while exposing intermediate temporal outputs, which enables PATA to learn a scene-wise inference time step.

\begin{figure*}[t]
    \centering
    \begin{minipage}{0.42\textwidth}
        \centering
        \includegraphics[width=\linewidth]{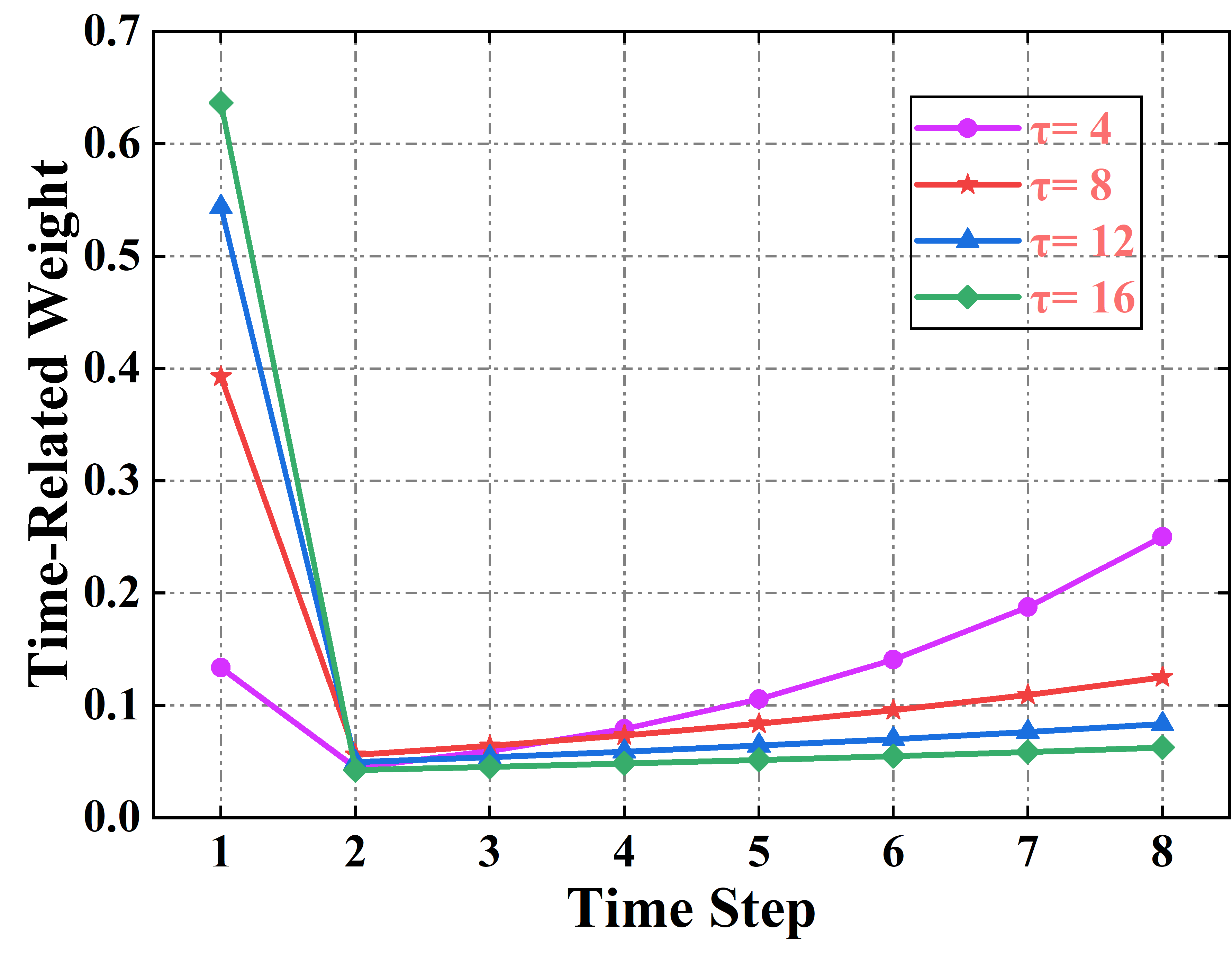}
        \subcaption{Per-step weight}
    \end{minipage}
    \begin{minipage}{0.42\textwidth}
        \centering
        \includegraphics[width=\linewidth]{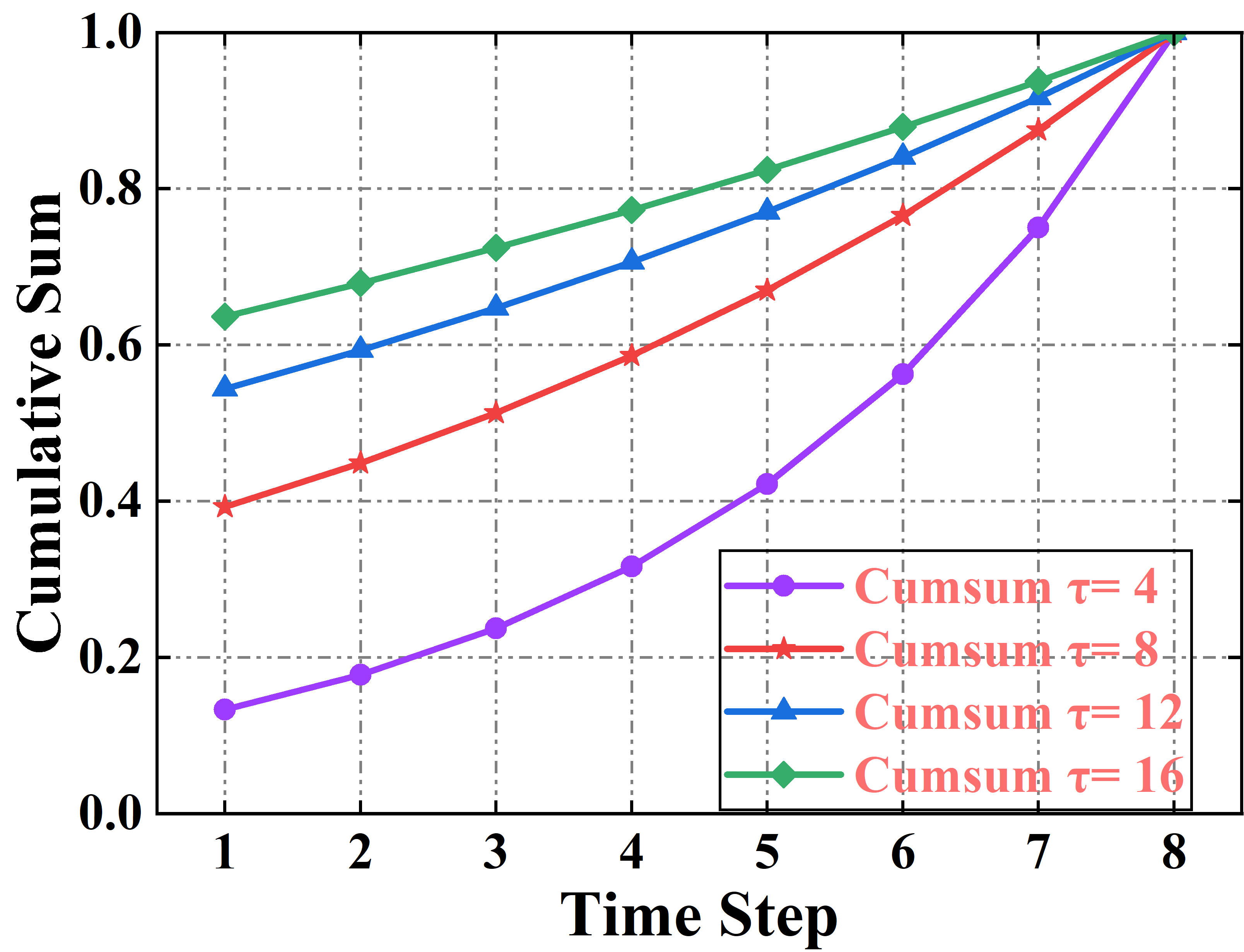}
        \subcaption{Cumulative weight}
    \end{minipage}
    \caption{Contribution distribution over time steps under the hybrid input mode. 
            (a) Per-step contribution weight. 
            (b) Cumulative contribution up to each time step. 
            The hybrid mode assigns a larger contribution to the first time step and reduces the dominance of late time steps.}
    \label{fig:time_weights}
\end{figure*}

\subsection{Hybrid Input Mode}
The effectiveness of temporal-budget learning depends on whether early time-step outputs already contain sufficient rendering information. In standard input modes, the accumulated membrane potential is often dominated by later time steps, which makes the rendered output sensitive to temporal truncation. We therefore analyze how different input modes distribute the output contribution across time steps and design a hybrid input mode that increases early-step contribution while retaining stable accumulation.
Given that our output layer accumulates membrane potentials without spike generation, we formulate the output at time step $T$ as:
\begin{equation}
    O_T = \sum_{t=0}^{T-1} w(t, \tau) \cdot \mathrm{func}(i_t)
\end{equation}
where $w(t, \tau)$ represents the contribution weight of time step $t$ to the final output, and $\mathrm{func}(\cdot)$ denotes the exponential function for the density network and sigmoid activation for the color network. 
For the density branch, the activation is applied to the density output, while the remaining hidden features $\mathbf{h}$ are passed to the color branch together with the viewing direction $\mathbf{d}$.
The weight function $w(t, \tau)$ follows distinct formulations depending on the input mode:
\begin{equation}
w(t, \tau) = 
\begin{cases}
    (1-\frac{1}{\tau})^{T-t-1} \frac{1}{\tau}, & \text{if decay input mode} \\
    (1-\frac{1}{\tau})^{T-t-1}, & \text{if non-decay input mode}
\end{cases}
\label{timeweight}
\end{equation}
We observe that both modes exhibit monotonically increasing weights toward later time steps, making the model performance sensitive to time step reduction. To address this, we propose a hybrid input scheme that strategically combines both modes:
\begin{equation}
v^l(t) = 
\begin{cases}
    i(t), & \text{if } t = 0 \\
    (1-\frac{1}{\tau})v^l(t-1) + \frac{1}{\tau}i(t), & \text{if } t > 0
\end{cases}
\label{hybridmode}
\end{equation}
This hybrid approach modifies the weight function as:
\begin{equation}
w(t, \tau) = 
\begin{cases}
    (1-\frac{1}{\tau})^{T-t-1}, & \text{if } t = 0 \\
    (1-\frac{1}{\tau})^{T-t-1} \frac{1}{\tau}, & \text{if } t > 0
\end{cases}
\label{hybridtimeweight}
\end{equation}
Crucially, this formulation maintains $\sum_{t=0}^{T-1} w(t, \tau) = 1$ for all $\tau > 1$. As demonstrated in Figure~\ref{fig:time_weights}, 
our hybrid mode significantly boosts the contribution of the initial time step while suppressing later time steps through appropriate $\tau$ selection. 
This mechanism enables effective time step reduction while reducing the sensitivity of rendering quality to temporal truncation.


\subsection{Pretraining-based Adaptive Time-step Adjustment}

\subsubsection{Overall Framework}
PATA treats the inference time step as a scene-wise temporal budget rather than a manually selected hyperparameter. 
This design follows the training and rendering structure of NeRF, where each scene has its own model and complexity, and all ray samples within the scene share a consistent time step to preserve parallel rendering.
We therefore parameterize the target time step $t^*$ as a learnable variable, distinct from the fixed maximum time step $T$:
\begin{equation}
    \begin{aligned}
    \hat{t}^*
    &= \mathrm{Clamp}(t^*, \mathrm{min}=1, \mathrm{max}=T),\\
    t_r^*
    &= \mathrm{Round}(\hat{t}^*)
    \end{aligned}
    \label{eq:time_step_quantization}
\end{equation}
where $\hat{t}^*$ denotes the constrained continuous training variable, while $t_r^*$ denotes
the discrete target time step used for inference. $\mathrm{Round}(\cdot)$ performs nearest-integer
rounding and $\mathrm{Clamp}(\cdot)$ maintains valid time-step bounds. To preserve gradient
propagation, we apply the straight-through estimator (STE) to the non-differentiable rounding
operation, where \(\frac{\partial \operatorname{Round}(x)}{\partial x}\) is approximated as 1
during backpropagation.

Directly optimizing $t^*$ from scratch couples two difficult processes: learning a reliable radiance field and reducing its temporal budget. This coupling makes the optimization prone to unstable early truncation. 
Our PATA framework separates these processes into two stages. The first stage learns a strong full-step spike-based renderer at the maximum time step $T$. 
The second stage activates the learnable target time step and compresses the temporal budget while keeping the target-step output aligned with the full-step prediction.

\subsubsection{Initial Stage Training}
The initial training phase establishes a robust pre-trained model for subsequent time-step optimization. We employ a composite loss function comprising three key components:
\begin{align}
    \mathcal{L}_{\mathrm{Render}}(T)
    &= M_T + \mathcal{L}_{\mathrm{consis}}(T) + \mathcal{L}_{\mathrm{Cauchy}} \label{render} \\
    \mathcal{L}_{\mathrm{consis}}(T)
    &= \gamma_0\|C_T - \mathrm{gt}\|_2 \notag \\
    &\quad + \gamma_1 \left(1 - \frac{C_T \cdot \mathrm{gt}}{\|C_T\| \|\mathrm{gt}\| + \epsilon}\right) \label{consis} \\
    \mathcal{L}_{\mathrm{Cauchy}}
    &= \gamma_2 \sum_{i} \log(1 + 2\sigma_i) \label{cauchy}
\end{align}
where $M_T$ denotes the mean squared error (MSE) between the rendered color $C_T$ (computed via Equation~\ref{pointscolor}) and ground truth $\mathrm{gt}$. The consistency loss $\mathcal{L}_{\mathrm{consis}}(T)$ combines L2 distance and cosine similarity to enforce both magnitude and directional alignment between rendered and ground truth images. Following \cite{shin2023binary}, the Cauchy regularization term $\mathcal{L}_{\mathrm{Cauchy}}$ promotes sparsity in the density predictions $\sigma_i$. The coefficients $\gamma_0$, $\gamma_1$, and $\gamma_2$ balance these loss components during optimization, which we set to $5 \times 10^{-5}$, $5 \times 10^{-2}$, and $1 \times 10^{-6}$, respectively.

\subsubsection{Temporal Budget Loss}
Following the initial training phase, we activate the trainable target time step $t^*$ as defined in Equation~\eqref{eq:time_step_quantization}. 
While our hybrid input mode enables control over time-step contributions through the decay parameter $\tau$, excessive $\tau$ values hinder effective training by limiting input information propagation (Equation~\eqref{hybridmode}). 
To address this, we introduce a temporal budget loss:
\begin{equation}
    \mathcal{L}_{\mathrm{budg}} = \alpha\mathcal{L}_{\mathrm{extra}} + \beta\mathcal{L}_{\mathrm{penalty}}
    \label{eq:adv_loss}
\end{equation}
Here $\alpha$ and $\beta$ are hyperparameters. The temporal budget loss creates a quality-aware competition between preserving necessary rendering information and reducing unnecessary late-step computation.
The first component $\mathcal{L}_{\mathrm{extra}}$ regulates both $\tau$ and $t^*$ through:
\begin{equation}
    \mathcal{L}_{\mathrm{extra}} = \frac{\|w_{i, t>t_r^*}\|_2}{\|W_{\mathrm{points}}\|_2} + \frac{\|c_{t>t_r^*}\|_2}{\|C_T\|_2} + e^{\tau}
    \label{eq:extra_loss}
\end{equation}
where $w_{i,t>t_r^*}$ represents weights for time steps beyond $t_r^*$, $W_{\mathrm{points}}$ denotes 
sampling point weights from Equation~\eqref{pointsweight}, $c_{t>t_r^*}$ is the weighted output 
for excess time steps, and $C_T$ is the final rendered color from Equation~\eqref{pointscolor}. 
The first two terms in ($\mathcal{L}_{\mathrm{extra}}$) penalize the remaining temporal 
contribution beyond $t_r^*$ from both the sampling weights and the rendered color output. 
If substantial information is still accumulated after $t_r^*$, directly truncating the 
inference at $t_r^*$ would cause noticeable rendering degradation. Therefore, 
$\mathcal{L}_{\mathrm{extra}}$ encourages the model to shift effective temporal 
contributions toward earlier time steps and preserves rendering quality under a reduced temporal budget. 
The exponential term $e^{\tau}$ prevents excessive growth of the decay factor $\tau$, avoiding 
degenerate solutions where the temporal weighting is dominated by an overly large membrane time constant.
To counterbalance $\mathcal{L}_{\mathrm{extra}}$'s tendency to increase $t^*$, we introduce a penalty term:
\begin{equation}
    \mathcal{L}_{\mathrm{penalty}} =
    \operatorname{sg}\left(\frac{M_T}{M_{\mathrm{smooth}}(\hat{t}^*)}\right)e^{\hat{t}^*}
    \label{eq:penalty}
\end{equation}

The penalty term supplies the compression force. 
Here $M_T$ is the full-step rendering error, and $M_{\mathrm{smooth}}(\hat{t}^*)$ is the continuous surrogate of the target-step error defined in Equation~\eqref{eq:smooth_mse}. 
The ratio $M_T/M_{\mathrm{smooth}}(\hat{t}^*)$ measures how close the target-step output is to the full-step output. 
When the target-step output approaches the full-step output, the penalty on $e^{\hat{t}^*}$ becomes stronger and pushes the model toward a smaller time step. 
When the target-step output is still inferior, the compression pressure is automatically weakened. 
Thus, the learned temporal budget is reduced only when the rendering quality can support the reduction.
The hyperparameter $\beta$ in Equation~\eqref{eq:adv_loss} controls the strength of 
this compression preference under different application requirements, and $\alpha$ is set 
to $1\times10^{-6}$ in our experiments.

\subsubsection{Dynamic Time-step Training}
While Equation~\eqref{eq:penalty} effectively links model performance to $t^*$, 
the rounding operation required to obtain $t_r^*$ introduces training instability. 
When $t^*$ fluctuates near integer midpoints, $t_r^*$ exhibits discontinuous jumps, causing abrupt changes in the target-step rendering error. 
To provide a continuous optimization signal for $t^*$, 
we linearly interpolate the rendering errors at the two adjacent integer time steps and apply the resulting smoothed error to both Equation~\eqref{render} and Equation~\eqref{eq:penalty}:
\begin{equation}
M_{\mathrm{smooth}}(\hat{t}^*)
=
\left(1-\delta\right) M_{t_f^*}
+
\delta M_{t_c^*},
\label{eq:smooth_mse}
\end{equation}
where $t_f^*=\lfloor \hat{t}^* \rfloor$, $t_c^*=\lceil \hat{t}^* \rceil$, and $\delta=\hat{t}^*-t_f^*$ denotes the fractional part of $\hat{t}^*$.
The smoothed loss $M_{\mathrm{smooth}}(\hat{t}^*)$ serves as a continuous surrogate for the discrete target-step rendering error. Instead of optimizing only the rounded output $C_{t_r^*}$, 
PATA evaluates the local quality landscape between the two adjacent integer time steps. 
This gives $t^*$ a stable optimization signal and prevents the training objective from changing abruptly when $t^*$ crosses a rounding boundary.

Since higher time steps typically provide more reliable SNN outputs~\cite{zuo2024self, QIU2024106475}, we employ temporal distillation from the full-step prediction. 
Specifically, we compute the MSE between outputs at $T$ and $t_r^*$:
\begin{equation}
    \mathcal{L}_{\mathrm{distill}} = \|C_T - C_{t_r^*}\|_2^2
    \label{eq:diss}
\end{equation}
At the beginning of the second training phase, the target-step output $C_{t_r^*}$ is weaker than the full-step output $C_T$, since the model has been pretrained with the maximum time step $T$. 
Thus, $\mathcal{L}_{\mathrm{distill}}$ uses $C_T$ as a temporal teacher signal to guide the lower-step output. 
This distillation does not require an external teacher model; the full-step output of the same network provides the supervision. 
Together with $\mathcal{L}_{\mathrm{extra}}$, it ensures that the information shifted to earlier time steps is aligned with the full-step rendering behavior. 
These objectives form a balanced optimization dynamic. 
$\mathcal{L}_{\mathrm{Render}}(\hat{t}^*)$ maintains direct supervision at the learned target step, 
$\mathcal{L}_{\mathrm{extra}}$ reduces the information tail after the inference cutoff, 
$\mathcal{L}_{\mathrm{penalty}}$ supplies quality-aware compression pressure, 
and $\mathcal{L}_{\mathrm{distill}}$ aligns the target-step output with the full-step prediction. 
As a result, the target time step is not reduced by an isolated regularizer; it is reduced only when the network has learned to render accurately within the shorter temporal budget.

The complete second-phase loss combines these components:
\begin{equation}
    \mathcal{L} = \mathcal{L}_{\mathrm{Render}}(\hat{t}^*) + \mathcal{L}_{\mathrm{budg}} + \mathcal{L}_{\mathrm{distill}}
    \label{eq:stageII_loss}
\end{equation}
The whole training process is summarized in Algorithm~\ref{alg:pata}. As a result, PATA converts time-step selection from a manually tuned hyperparameter into a learned scene-wise budget. 
At inference time, the model is unrolled only to $t_r^*$, directly reducing temporal computation while preserving the rendering behavior learned from the full-step model.

\begin{algorithm}[t]
\caption{Pretraining-based Adaptive Time-step Adjustment (PATA)}
\label{alg:pata}
\begin{algorithmic}[1]
\Require Training rays and ground-truth colors $\mathcal{D}$, maximum time step $T$, first-stage iterations $N_1$, second-stage iterations $N_2$, and loss weights $\alpha$ and $\beta$.
\Ensure Optimized network parameters $\Theta$, PLIF decay factors $\tau$, and scene-wise target time step $t_r^*$.

\State Initialize SNN-NeRF parameters $\Theta$ and PLIF decay factors $\tau$.

\For{$k = 1$ to $N_1$}
\State Sample a mini-batch $\mathcal{B}$ from $\mathcal{D}$.
\State Unroll the spike-based NeRF for $T$ time steps and render the full-step output $C_T$.
\State Compute the first-stage rendering loss $\mathcal{L}_{\mathrm{Render}}(T)$.
\State Update $\Theta$ and $\tau$ by backpropagation.
\EndFor

\State Initialize the learnable target time step $t^* \leftarrow T-1$.

\For{$k = 1$ to $N_2$}
\State Sample a mini-batch $\mathcal{B}$ from $\mathcal{D}$.
\State Compute $\hat{t}^*=\mathrm{Clamp}(t^*,1,T)$ and $t_r^*=\mathrm{Round}(\hat{t}^*)$.
\State Compute $t_f^*=\lfloor\hat{t}^*\rfloor$, $t_c^*=\lceil\hat{t}^*\rceil$, and $\delta=\hat{t}^*-t_f^*$.
\State Unroll the spike-based NeRF for $T$ time steps and obtain $C_{t_f^*}$, $C_{t_c^*}$, $C_{t_r^*}$, and $C_T$.
\State Compute $M_{\mathrm{smooth}}(\hat{t}^*)=(1-\delta)M_{t_f^*}+\delta M_{t_c^*}$.
\State Compute the smoothed rendering loss $\mathcal{L}_{\mathrm{Render}}(\hat{t}^*)$ by replacing $M_T$ in Equation~\eqref{render} with $M_{\mathrm{smooth}}(\hat{t}^*)$.
\State Compute $\mathcal{L}_{\mathrm{penalty}}=\operatorname{sg}\left(M_T/M_{\mathrm{smooth}}(\hat{t}^*)\right)e^{\hat{t}^*}$.
\State Compute the temporal budget loss $\mathcal{L}_{\mathrm{budg}}=\alpha\mathcal{L}_{\mathrm{extra}}+\beta\mathcal{L}_{\mathrm{penalty}}$.
\State Compute the temporal distillation loss $\mathcal{L}_{\mathrm{distill}}=|C_T-C_{t_r^*}|_2^2$.
\State Compute the second-stage objective $\mathcal{L}=\mathcal{L}_{\mathrm{Render}}(\hat{t}^*)+\mathcal{L}_{\mathrm{budg}}+\mathcal{L}_{\mathrm{distill}}$.
\State Update $\Theta$, $\tau$, and $t^*$ by backpropagation.
\EndFor

\State Fix the learned $t_r^*$ as the scene-wise inference time step.
\State During inference, unroll the spike-based NeRF only for $t_r^*$ time steps and render $C_{t_r^*}$.
\end{algorithmic}
\end{algorithm}


\subsection{Max Time-Step Scaling}

Although PATA reduces the inference time step to the learned $t_r^*$, the second training stage still unrolls the network to the maximum time step $T$. This full unrolling is necessary for computing the full-step teacher output, adjacent-step rendering errors, and temporal budget loss. However, early NeRF training often contains many unfiltered sampling points before the occupancy or importance sampling mechanism becomes stable, which amplifies the memory cost of full-time-step unrolling.

To keep training stable under large $T$, we adopt a max time-step scaling strategy. At the beginning of training, the model is optimized with a reduced maximum time step, which avoids the peak memory cost caused by dense early samples. Once the sampling process becomes stable and irrelevant points are filtered out, we restore the maximum time step to $T$ and train the full PATA objective. This strategy preserves the full temporal capacity required by PATA while removing the initial memory bottleneck.

\section{RESULTS}




\subsection{Experimental Setup}

\textbf{Datasets and metrics.}
We evaluate PATA on standard novel-view synthesis benchmarks. For the INGP-NeRF backbone, we use the Synthetic-NeRF dataset~\cite{mildenhall2021nerf} and the Mip-NeRF 360 dataset~\cite{barron2022mip}. For the TensoRF backbone, we use the Synthetic-NeRF dataset and the LLFF forward-facing scenes~\cite{mildenhall2019local}, following the common evaluation setting of tensor-factorized NeRF models. Rendering quality is measured by PSNR and SSIM. Efficiency is reported by the learned inference time step and estimated energy consumption.

\textbf{Backbones.}
We instantiate PATA on two efficient NeRF backbones with different scene representations. INGP-NeRF uses a multiresolution hash-grid representation followed by lightweight density and color networks. TensoRF represents the radiance field with tensor-factorized density and appearance components, and decodes the appearance feature into RGB values through a color network. These two backbones provide complementary validation settings: one based on hash-grid encoding and one based on tensor factorization.

\textbf{INGP-NeRF implementation.}
For INGP-NeRF, our implementation is built upon the $torch$\textendash$ngp$ framework~\cite{torch-ngp}. The density network contains one hidden layer and the color network contains two hidden layers, both with 128 hidden units. We replace the nonlinear activations in both prediction networks with PLIF neurons and use output PLIF layers to accumulate temporal estimates of density $\sigma$ and color $\mathbf{c}$. The maximum time step is set to $T=8$, and the target time step is initialized as $t^*=7$ after the pretraining phase.

\textbf{TensoRF implementation.}
For TensoRF, we keep the tensor-factorized density field, appearance field, sampling procedure, and volume rendering process unchanged. Different from INGP-NeRF, TensoRF does not use an MLP density network; density is computed from the factorized density components, while the appearance feature is decoded by a color network. Therefore, PATA is applied to the color decoder of TensoRF. Specifically, we replace the original MLP color decoder with a spike-based feature MLP color decoder. The decoder has two hidden layers with 128 channels and PLIF neurons between linear layers, followed by a linear RGB output layer. The maximum time step is set to $T=8$, the initial membrane time constant is $\tau=2.0$, and the firing threshold is set to 0.5.

For both Synthetic-NeRF and LLFF, we follow the original TensoRF configuration for the tensor-factorized field and rendering pipeline. Under these settings, the spike-based color decoder receives a 150-dimensional input on Synthetic-NeRF and a 30-dimensional input on LLFF.

\textbf{Training details.}
All experiments are implemented in PyTorch. The spike-based modules are built with SpikingJelly~\cite{fang2023spikingjelly}, and all models are trained on a single NVIDIA RTX 5880 GPU with mixed precision. Unless otherwise specified, PATA follows the same two-stage training protocol for both INGP-NeRF and TensoRF. The first stage trains a fixed full-time-step spike-based renderer with the maximum time step $T=8$ for 10,000 iterations. The second stage starts from the first-stage checkpoint and optimizes the learnable target time step together with the rendering objective for 30,000 iterations.

At the beginning of each training stage, the maximum time step is set to 2 for the initial epoch and then restored to 8, following the max time-step scaling strategy. We report only the adaptive second-stage results for the TensoRF experiments.


\subsection{Quantitative Comparison}
\subsubsection{Quality-Efficiency Trade-off on INGP-NeRF}
\label{sec:exp}

\begin{table*}[!t]
    \caption{Quality-efficiency trade-off of PATA on the INGP-NeRF backbone. The values in parentheses denote the energy reduction relative to the original INGP-NeRF model, and the best result is highlighted in bold.}
    \centering
    \small
    \resizebox{\linewidth}{!}{
    \begin{tabular}{cc cccc cccc}
     \toprule
            \multicolumn{2}{c}{Dataset} & \multicolumn{4}{c}{Synthetic-NeRF} & \multicolumn{4}{c}{Mip-NeRF 360} \\
            \cmidrule(lr){3-6}\cmidrule(lr){7-10}
            \multicolumn{2}{c}{Method} & PSNR$\uparrow$ & SSIM$\uparrow$ & Time Step$\downarrow$ &  Energy(mJ)$\downarrow$ & PSNR$\uparrow$ & SSIM$\uparrow$ & Time Step$\downarrow$ & Energy(mJ) $\downarrow$ \\
     \midrule
            \multicolumn{2}{c}{ANN-INGP-NeRF}        & 32.14 & \textbf{0.959} & \_ & 955.27  & \textbf{25.48} & \textbf{0.668} & \_ & $1.898 \times 10^{4}$  \\
            \midrule
            \multirow{3}*{Ours} & $\beta = 1 \times 10^{-7}$ & \textbf{32.45} & \textbf{0.959} & 5.125 & 643.39(32.65\%) & 25.30 & 0.667 & 6.000 &  $1.560 \times 10^{4}$(17.85\%)  \\  
            
            & $\beta = 5 \times 10^{-7}$ & 32.21 & 0.957 & 3.625 &  492.25(48.47\%) & 25.09 & 0.646 & 4.857 & $1.366 \times 10^{4}$(28.02\%)  \\

            & $\beta = 1 \times 10^{-6}$ & 32.04 & 0.956 & 2.875 & 405.33(\textbf{57.57\%}) & 25.01 & 0.641 & 4.000 & $1.187 \times 10^{4}$(\textbf{37.46\%}) \\
    \bottomrule
    \end{tabular}
    }
    \label{tab:psrnandenergy}
\end{table*}

Table~\ref{tab:psrnandenergy} reports the quality-efficiency trade-off of PATA on the INGP-NeRF backbone. By varying the temporal budget weight $\beta$ in Equation~\eqref{eq:adv_loss}, PATA learns different scene-wise inference time steps while preserving competitive rendering quality. On Synthetic-NeRF, PATA reduces the average time step from the full budget of $T=8$ to 5.125, 3.625, and 2.875 under increasing $\beta$, corresponding to 32.65\%, 48.47\%, and 57.57\% estimated energy reduction, respectively. On Mip-NeRF 360, the same trend is observed: the learned time step decreases from 6.000 to 4.000, yielding up to 37.46\% estimated energy reduction.

The three $\beta$ settings separate high-fidelity and efficiency-oriented regimes. At $\beta=1\times10^{-7}$, PATA reaches 32.45 dB PSNR on Synthetic-NeRF with 32.65\% estimated energy reduction. Increasing $\beta$ to $1\times10^{-6}$ reduces the average time step to 2.875 and the estimated energy to 405.33mJ, with PSNR remaining at 32.04 dB. The same setting also gives the largest Mip-NeRF 360 energy reduction in this table, reducing the estimated energy from $1.898\times10^{4}$mJ to $1.187\times10^{4}$mJ.

\subsubsection{Backbone Generalization on TensoRF}

We further instantiate PATA on TensoRF to evaluate temporal-budget learning beyond the hash-grid backbone. This setting differs structurally from INGP-NeRF, as TensoRF computes density from tensor-factorized components and uses a color decoder only for appearance prediction. PATA is therefore applied to the TensoRF color decoder while preserving the tensor-factorized radiance-field representation and the original volume rendering pipeline.

\begin{table*}[!t]
    \caption{Quality-efficiency trade-off of PATA on the TensoRF backbone. The values in parentheses denote the energy reduction relative to the ANN-TensoRF baseline, and the best result is highlighted in bold.}
    \centering
    \fontsize{10}{\baselineskip}\selectfont
    \resizebox{\linewidth}{!}{
    \begin{tabular}{cc cccc cccc}
     \toprule
            \multicolumn{2}{c}{Dataset} & \multicolumn{4}{c}{Synthetic-NeRF} & \multicolumn{4}{c}{LLFF} \\
            \cmidrule(lr){3-6}\cmidrule(lr){7-10}
            \multicolumn{2}{c}{Method} & PSNR$\uparrow$ & SSIM$\uparrow$ & Time Step$\downarrow$ & Energy(mJ)$\downarrow$ & PSNR$\uparrow$ & SSIM$\uparrow$ & Time Step$\downarrow$ & Energy(mJ)$\downarrow$ \\
     \midrule
            \multicolumn{2}{c}{ANN-TensoRF} & 33.14 & 0.963 & -- & $1.261 \times 10^{3}$ & 26.73 & 0.839 & -- & $1.191 \times 10^{4}$ \\
            \midrule
            \multirow{3}*{Ours}

            & $\beta = 5 \times 10^{-9}$ & \textbf{33.10} & \textbf{0.961} & 5.625 & 885.2 (\textbf{29.81\%}) & \textbf{26.62} & \textbf{0.832} & 5.625 & $6.341 \times 10^{3}$(\textbf{46.76\%}) \\

            & $\beta = 5 \times 10^{-8}$ & 33.04 & 0.960 & 3.875 & 815.2(35.36\%) & 26.58 & 0.831 & 4.125 & $5.314 \times 10^{3}$(55.39\%) \\

            & $\beta = 5 \times 10^{-7}$ & 32.91 & 0.959 & 2.000 & 748.4(40.66\%) & 26.50 & 0.828 & 1.875 & $3.704 \times 10^{3}$(68.90\%) \\     
    \bottomrule
    \end{tabular}
    }
    \label{tab:tensorf_quality_energy}
\end{table*}

Table~\ref{tab:tensorf_quality_energy} reports the quality-efficiency trade-off on the TensoRF backbone. On Synthetic-NeRF, the high-fidelity setting with $\beta=5\times10^{-9}$ achieves 33.10 dB PSNR and 0.961 SSIM, closely matching ANN-TensoRF while reducing the estimated energy by 29.81\%. As $\beta$ increases, the learned time step decreases from 5.625 to 2.000, and the estimated energy reduction increases to 40.66\%, with PSNR changing from 33.10 dB to 32.91 dB. On LLFF, the estimated energy reduction reaches 46.76\% at the high-fidelity setting and 68.90\% under the efficiency-oriented setting, with PSNR ranging from 26.50 dB to 26.62 dB.

The larger energy reduction on LLFF comes from the different TensoRF decoder configuration. For Synthetic-NeRF, the TensoRF color decoder uses a 150-dimensional input, so the dense decoder projection and the remaining tensor-factorized rendering computation retain a larger FLOP contribution in the overall energy estimate. For LLFF, the decoder input dimension is only 30; under this configuration, the MLP portion replaced by the spike-based color decoder occupies a larger share of the reducible computation relative to the remaining rendering pipeline. Reducing the learned color-decoder time step therefore changes a larger portion of the total estimated energy on LLFF.


\subsubsection{Comparison with Spike-based NeRF Models}

We compare PATA with existing spike-based NeRF methods on the Synthetic-NeRF dataset, where most prior spike-based neural rendering models report their results. Since PATA can be instantiated on different rendering backbones, Table~\ref{tab:spikenerf} reports representative INGP-NeRF and TensoRF variants selected from the quality-efficiency trade-offs above.

\begin{table}[t]
\centering
\fontsize{10}{\baselineskip}\selectfont
    \caption{Comparison with spike-based NeRF models on the Synthetic-NeRF dataset. The best result is highlighted in bold.}
    \label{tab:spikenerf}
    \resizebox{\linewidth}{!}{
    \begin{tabular}{lccc}
        \toprule
        Method  & PSNR$\uparrow$ & SSIM$\uparrow$ & Time Step$\downarrow$   \\
        \midrule
        Spiking-NeRF \cite{li2025spiking}       & 30.41  & \_    & 256 \\
        Mixed Spiking-NeRF \cite{wang2025mixed} & 31.61  & 0.949 & 4 \\
        SpikingNeRF-D \cite{yao2023spikingnerf}      & 31.64  & 0.952 & 4 \\
        SpikingNeRF-T \cite{yao2023spikingnerf}      & 32.45  & 0.956 & \textbf{1} \\
        Spik-NeRF-Wan \cite{wan2026spik} & 30.23  & 0.942 & 2 \\
        Ours-INGP-NeRF           & 32.21  & 0.957 & 3.625 \\
        Ours-TensoRF             & \textbf{32.91}  & \textbf{0.959} & 2 \\
        \bottomrule
    \end{tabular}
    }
\end{table}

Table~\ref{tab:spikenerf} shows that PATA achieves the strongest rendering quality among the compared spike-based NeRF models. The INGP-NeRF variant reaches 32.21 dB PSNR and 0.957 SSIM with an adaptive average time step of 3.625. The TensoRF variant further improves the average PSNR to 32.91 dB and SSIM to 0.959, surpassing SpikingNeRF-T in rendering quality while using an average time step of 2.

\subsubsection{Comparison with ANN-based NeRF Models}

We further compare PATA with representative ANN-based NeRF methods under standard neural rendering benchmarks. Table~\ref{tab:anns} reports the corresponding PATA variants according to the evaluated backbone and dataset. 

\begin{table*}[t]
    \centering
    \small
    \setlength{\tabcolsep}{3.5pt}
    \caption{Comparison with ANN-based NeRF methods on Synthetic-NeRF, Mip-NeRF 360, and LLFF. The best results on each dataset are shown in bold.}
    \label{tab:anns}
    \resizebox{\linewidth}{!}{ 
    \begin{tabular}{lcc lcc lcc}
            \toprule
            \multicolumn{3}{c}{Synthetic-NeRF} & \multicolumn{3}{c}{Mip-NeRF 360} & \multicolumn{3}{c}{LLFF} \\
            \cmidrule(lr){1-3}\cmidrule(lr){4-6}\cmidrule(lr){7-9}
            Method  & PSNR$\uparrow$ & SSIM$\uparrow$ & Method & PSNR$\uparrow$ & SSIM$\uparrow$ & Method & PSNR$\uparrow$ & SSIM$\uparrow$   \\
            \midrule
            JaxNeRF\cite{jaxnerf2020github}   & 31.65  & 0.952     &  NeRF\cite{mildenhall2021nerf}         & 23.85  & 0.605 & SRN\cite{sitzmann2019scene} & 22.84 & 0.668 \\
            MipNeRF\cite{barron2021mip}  & 33.09  & 0.961     &  NeRF++\cite{zhang2020nerf++}       & 25.11  & 0.676 & LLFF\cite{mildenhall2019local} & 24.13 & 0.798 \\
            TensoRF\cite{tensorf}   & \textbf{33.14}  & \textbf{0.963}     &  MipNeRF\cite{barron2021mip}     & 24.04  & 0.616 & NeRF\cite{mildenhall2021nerf} & 26.50 & 0.811 \\
            INGP\cite{muller2022instant} & 32.14  & 0.959     &  INGP\cite{muller2022instant}    & 25.48  & 0.668 & MipNeRF\cite{barron2021mip} & 26.60 & 0.817 \\
            SRes\cite{dai2023sres} & 32.52  & 0.959     &  Nerfacto\cite{tancik2023nerfstudio}     & 26.39  & 0.731 & Plenoxels\cite{plenoxels} & 26.29 & \textbf{0.839} \\
            BiRF\cite{shin2023binary}      & 32.64  & 0.958     &  ZipNeRF\cite{barron2023zip}     & \textbf{28.54}  & \textbf{0.828} & TensoRF\cite{tensorf} & \textbf{26.73} & \textbf{0.839} \\
            \midrule
            Ours-INGP-NeRF  & 32.21  & 0.957     &  Ours-INGP-NeRF          & 25.09  & 0.646 & -- & -- & -- \\
            Ours-TensoRF  & 33.10  & 0.961     &  --          & --  & -- & Ours-TensoRF & 26.62 & 0.832 \\
            \bottomrule
    \end{tabular}
    }
\end{table*}

Table~\ref{tab:anns} shows that PATA preserves competitive rendering quality compared with ANN-based NeRF models. On Synthetic-NeRF, the high-fidelity Ours-TensoRF setting reaches 33.10 dB PSNR and 0.961 SSIM, closely approaching the ANN TensoRF result while using adaptive spike-based temporal computation. Ours-INGP-NeRF also remains close to the original INGP-NeRF backbone, with the energy reduction shown in Table~\ref{tab:psrnandenergy}. On LLFF, Ours-TensoRF achieves 26.62 dB PSNR and 0.832 SSIM, matching the quality range of representative forward-facing NeRF renderers while reducing the estimated energy by 46.76\% as reported in Table~\ref{tab:tensorf_quality_energy}. On Mip-NeRF 360, Ours-INGP-NeRF achieves comparable quality to several conventional ANN baselines, while high-capacity ANN renderers such as ZipNeRF still achieve higher reconstruction accuracy.

\subsubsection{Energy Efficiency Analysis}

Following the standard operation-level energy estimation protocol used in SNN studies~\cite{shi2024spikingresformer,yao2023spikingnerf}, we estimate inference energy from floating-point operations and spike-based operations under 45nm technology:
\begin{gather}
    SOPs = fr \times T \times FLOPs,\\
    Energy_{SOPs} = 0.9pJ \times SOPs,\\
    Energy_{FLOPs} = 4.6pJ \times FLOPs,\\
    Energy_{total} = Energy_{SOPs} + Energy_{FLOPs},
\end{gather}
where $fr$ denotes the average firing rate, $T$ denotes the inference time step, $SOPs$ denotes spike-based operations, and $FLOPs$ denotes floating-point operations. The energy costs of one spike-based operation and one floating-point operation are set to $0.9pJ$ and $4.6pJ$, respectively. This protocol follows the common evaluation practice of prior SNN and spike-based NeRF work and reports the operation-level energy of spike-based inference.

Tables~\ref{tab:psrnandenergy} and~\ref{tab:tensorf_quality_energy} summarize the estimated energy results across the two evaluated backbones. On INGP-NeRF, PATA reduces the average time step from the full budget of $T=8$ and achieves up to \textbf{57.57\%} energy reduction on Synthetic-NeRF and \textbf{37.46\%} energy reduction on Mip-NeRF 360 while maintaining competitive PSNR and SSIM. On TensoRF, the estimated energy reduction reaches \textbf{40.66\%} on Synthetic-NeRF and \textbf{68.90\%} on LLFF, with the tensor-factorized density field, appearance field, sampling procedure, and volume rendering pipeline kept unchanged. The stronger reduction on LLFF is consistent with the smaller 30-dimensional TensoRF color-decoder input in this setting, where the spike-based decoder accounts for a larger share of the reducible computation.

The reported energy values follow an operation-level estimate for neuromorphic or customized low-power accelerators, where sparse spike operations can be exploited, such as FPGA- and ASIC-based accelerators~\cite{KARAMIMANESH2025107256, DENG202331, longchar2024spalen}. This metric quantifies the inference-energy effect of temporal-budget learning under the same assumptions used in prior SNN evaluations.

\subsection{Qualitative Results}

Figure~\ref{fig:render} presents qualitative results on the INGP-NeRF backbone. Under reduced target time steps, PATA preserves the main object geometry, texture appearance, and background structure across representative Synthetic-NeRF and Mip-NeRF 360 scenes. These visual results are aligned with the quantitative trends in Table~\ref{tab:psrnandenergy}, where the learned temporal budgets reduce inference cost while maintaining rendering quality.

\begin{center}
    \includegraphics[width=1.11\columnwidth]{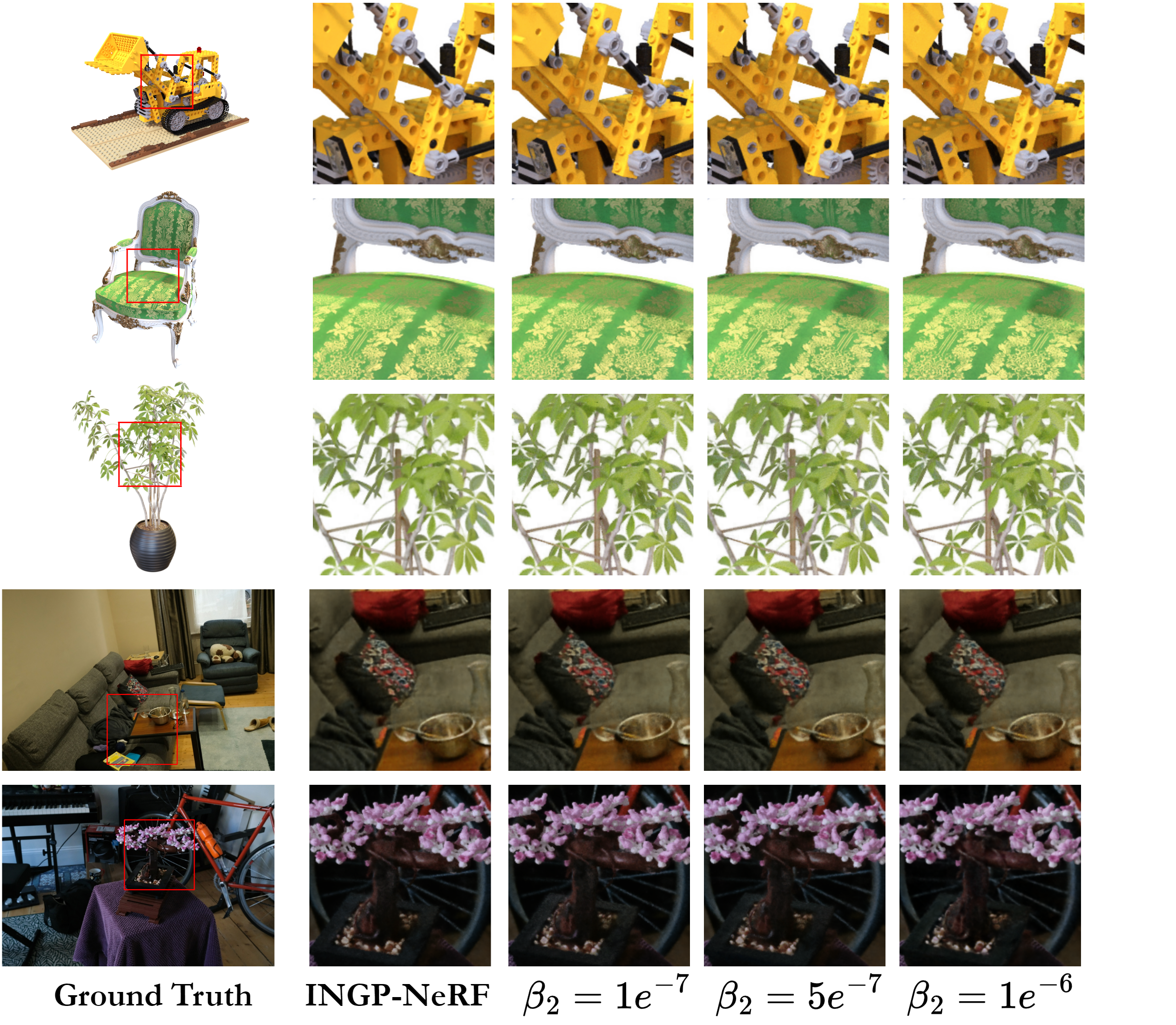}
    \captionof{figure}{\textbf{Qualitative results} on representative Synthetic-NeRF and Mip-NeRF 360 scenes using the INGP-NeRF backbone.}
    \label{fig:render}
\end{center}

Figure~\ref{fig:tensorf_llff_render} further compares GT images, ANN-TensoRF renderings, and PATA-TensoRF renderings on LLFF. In this setting, PATA replaces only the TensoRF color decoder with the spike-based adaptive decoder, while the tensor-factorized density field, appearance field, sampling process, and volume rendering pipeline remain unchanged. With $\beta=5\times10^{-7}$, PATA-TensoRF preserves the scene layout and local visual details across real forward-facing scenes, matching the small PSNR/SSIM changes reported in Table~\ref{tab:tensorf_quality_energy}. This comparison provides qualitative evidence that the proposed temporal-budget learning can be applied beyond the hash-grid INGP-NeRF backbone.

\begin{center}
    \includegraphics[width=0.95\columnwidth]{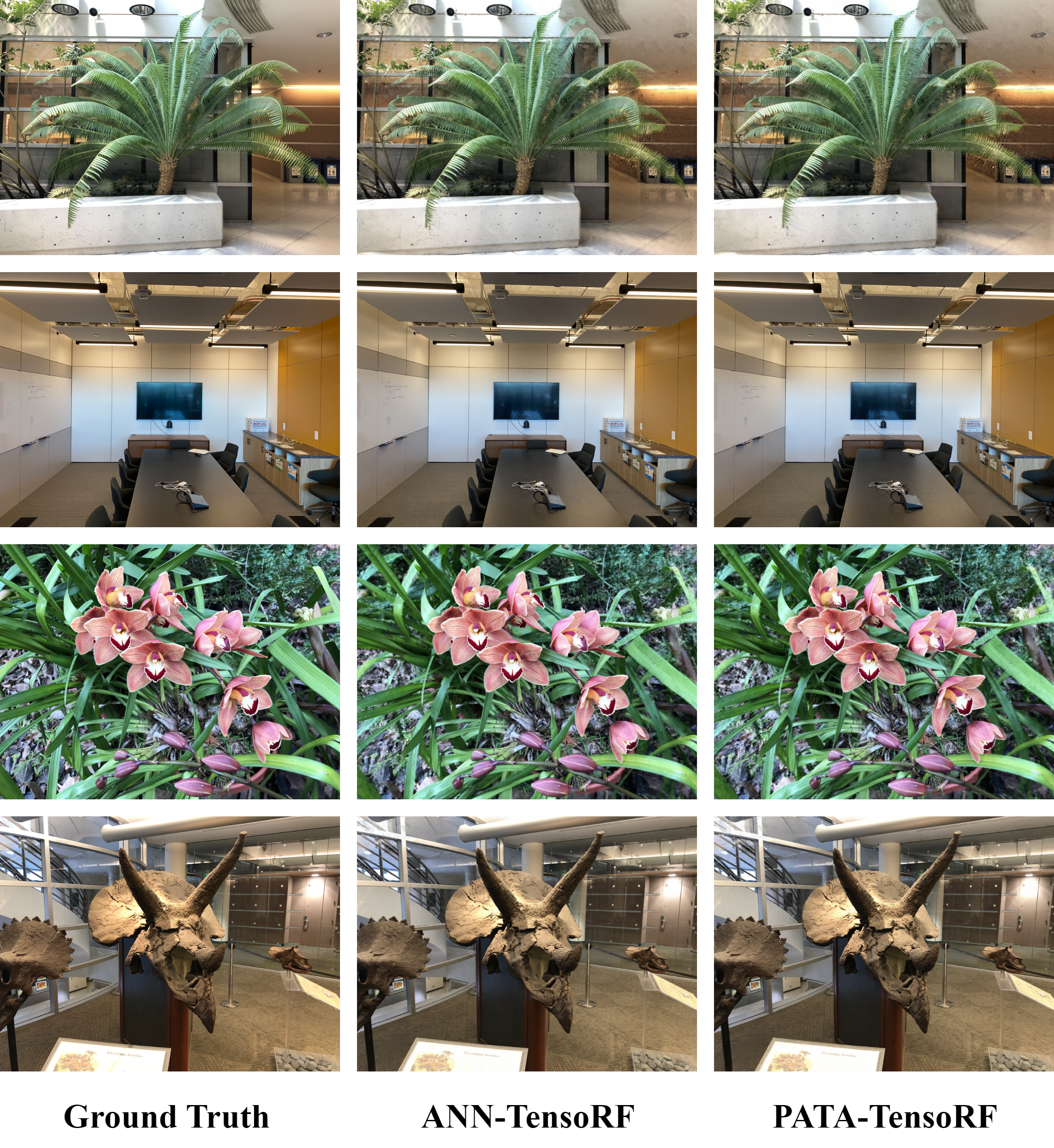}
    \captionof{figure}{\textbf{Qualitative comparison on LLFF using the TensoRF backbone.} Columns show GT, ANN-TensoRF, and PATA-TensoRF ($\beta=5\times10^{-7}$), respectively.}
    \label{fig:tensorf_llff_render}
\end{center}

Figure~\ref{fig:three-subfigures} analyzes the learned time-step allocation across scenes using the INGP-NeRF backbone under the same parameter setting as Figure~\ref{fig:tensorf_llff_render}. Scenes with simpler textures, such as chair, hotdog, and room, use fewer time steps, while scenes with more complex geometry or texture, such as drums and bicycle, receive larger temporal budgets. Since all sampling points within a scene share the same target time step, this scene-level allocation preserves parallel inference while adapting computation to scene complexity.

\begin{center}
    \includegraphics[width=0.96\columnwidth]{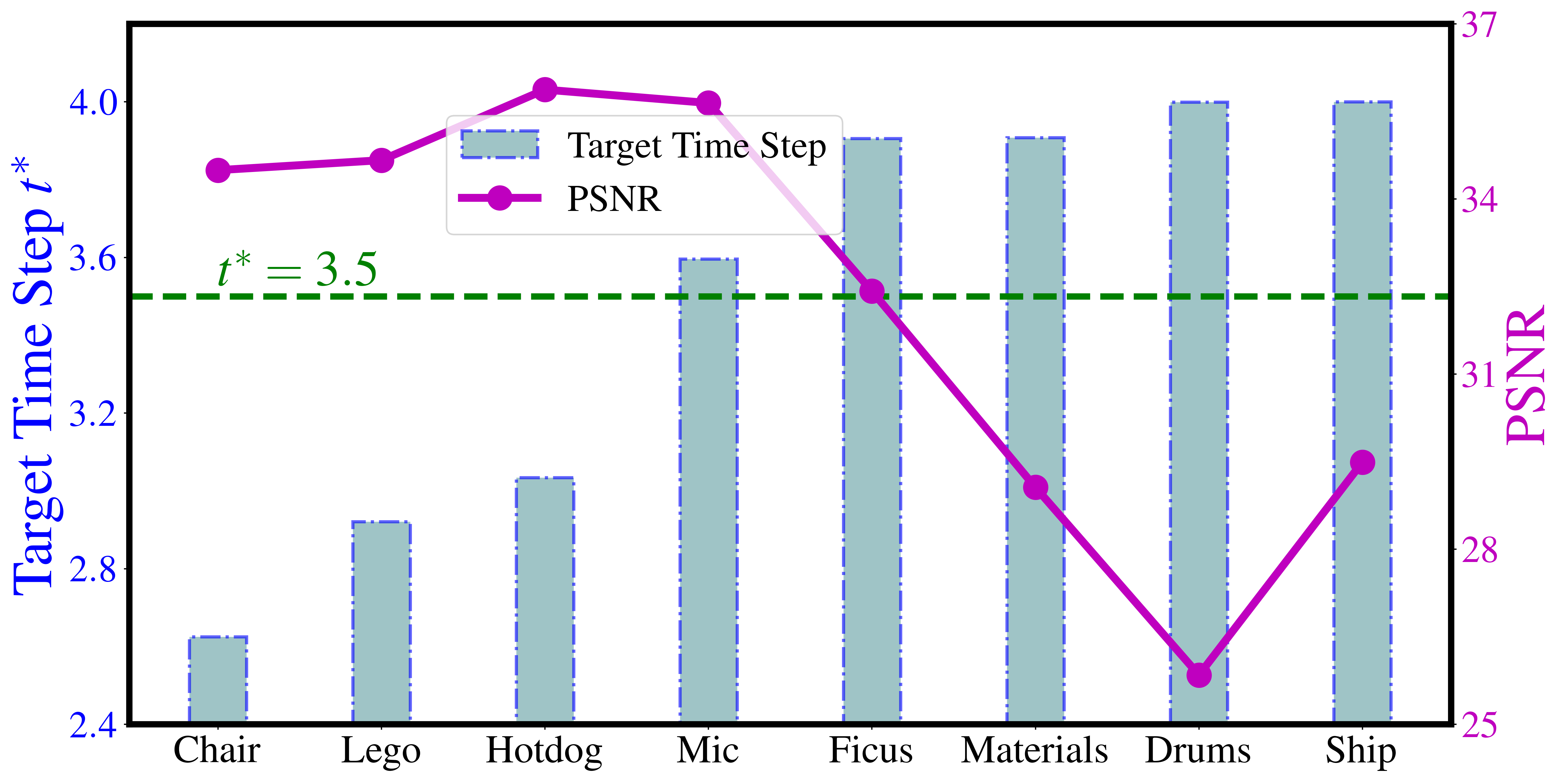}\\[-0.4em]
    {\footnotesize (a) Synthetic-NeRF}\\[0.4em]
    \includegraphics[width=0.96\columnwidth]{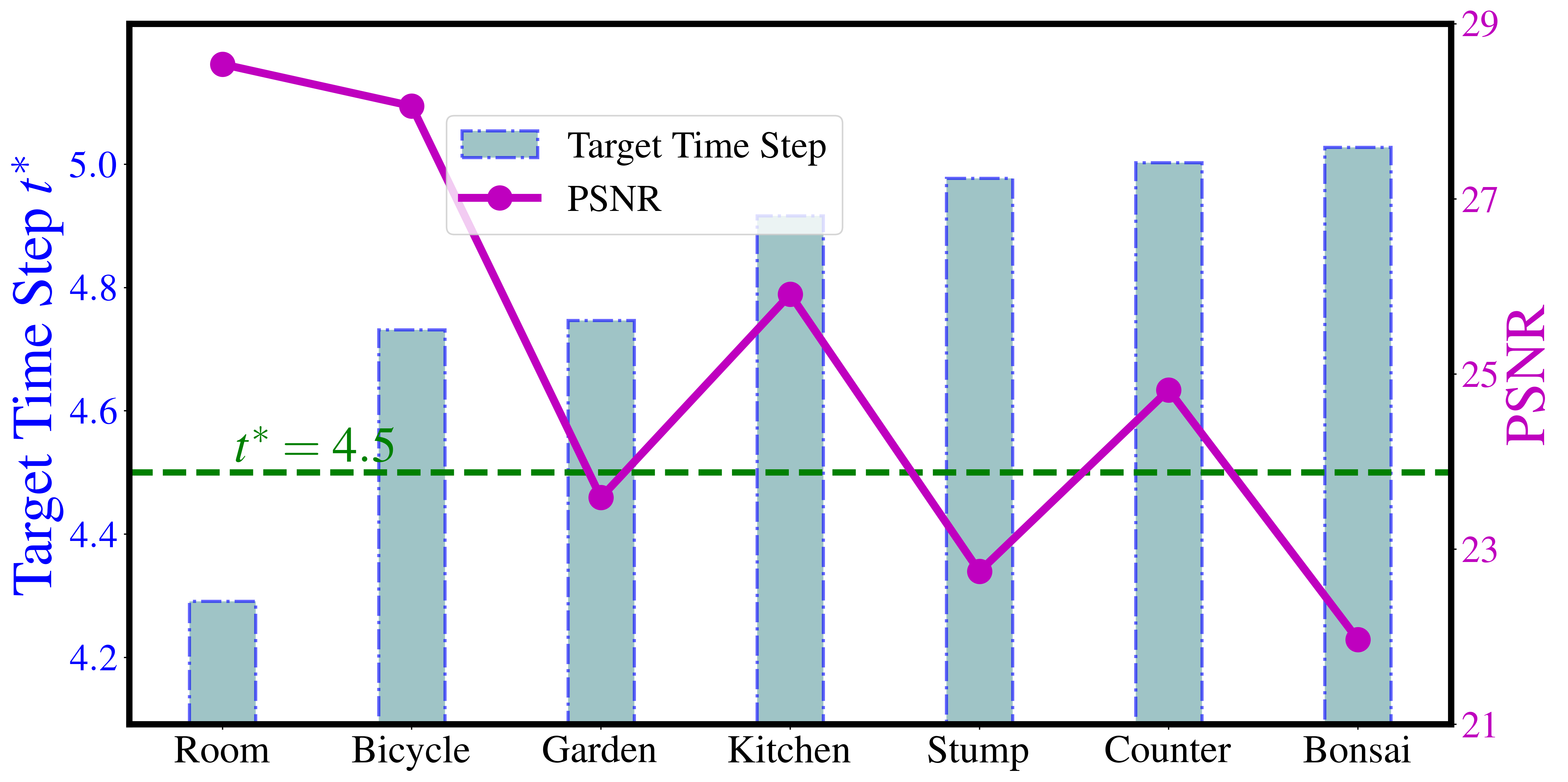}\\[-0.4em]
    {\footnotesize (b) Mip-NeRF 360}
    \captionof{figure}{Relationship between PSNR and target time steps for different scenes.}
    \label{fig:three-subfigures}
\end{center}

\subsection{Ablation Study}
We conduct ablation studies on the Synthetic-NeRF dataset using the INGP-NeRF backbone under the efficiency-oriented setting with $\beta=5\times10^{-7}$, unless otherwise specified. 
These experiments examine four design choices in PATA: the hybrid input mode, the two-stage training strategy, the temporal budget loss, and the learned target time step.

\begin{table}[h]
\centering
    \caption{Ablation study of input modes on Synthetic-NeRF using the INGP-NeRF backbone. The best result is highlighted in bold.}
    \begin{tabular}{lccc}
            \toprule
            Method  & PSNR$\uparrow$ & SSIM$\uparrow$ & Time Step$\downarrow$   \\
            \midrule

            Decay          & 31.82  & 0.955              &   8.000\\
            No-Decay       & 32.09  & 0.956              &   8.000\\
            Hybrid       & \textbf{32.12}  & \textbf{0.957}     &   8.000\\
            \bottomrule
    \end{tabular}
    \label{tab:decaymode}
\end{table}


Table~\ref{tab:decaymode} evaluates the input mode used in the spike-based renderer. The decay mode scales the input contribution at each time step by $1/\tau$, 
which weakens membrane accumulation and leads to the lowest PSNR. The no-decay mode avoids this attenuation and improves PSNR, but it does not provide explicit 
control over the temporal contribution of different steps. The hybrid mode combines stable accumulation with controllable temporal weighting and achieves the best 
PSNR and SSIM among the three settings. We therefore use the hybrid mode as the default input mechanism before optimizing the temporal budget in the second stage.

Table~\ref{tab:twostage} evaluates the two-stage training strategy. The hybrid baseline trains a fixed full-step spike-based renderer with $T=8$, 
while the w/o pretrain setting directly optimizes the adaptive target time step from initialization. Direct adaptation reaches the same average time step as PATA but reduces PSNR by 0.62 dB. 
Starting from a full-step pretrained model provides a stronger rendering initialization, allowing the second stage to reduce the temporal budget while preserving higher reconstruction quality.

\begin{table}[h]
\centering
    \caption{Ablation study of two-stage training. The best result is highlighted in bold.}
    \begin{tabular}{lccc}
            \toprule
            Method  & PSNR$\uparrow$ & SSIM$\uparrow$ & Time Step$\downarrow$   \\
            \midrule
            Hybrid       & 32.12  & \textbf{0.957}     &   8.000\\
            w/o pretrain   & 31.59  & 0.954     &   \textbf{3.625}\\
            Ours           & \textbf{32.21}  & \textbf{0.957}     &   \textbf{3.625}\\
            \bottomrule
    \end{tabular}
    
    \label{tab:twostage}
\end{table}

\begin{table}[h]
\centering
    \caption{Ablation study of temporal budget loss components. The best result is highlighted in bold.}
    \begin{tabular}{lccc}
            \toprule
            Method  & PSNR$\uparrow$ & SSIM$\uparrow$ & Time Step$\downarrow$   \\
            \midrule
            w/o $\mathcal{L}_{extra}$   & 31.69  & 0.953     &   \textbf{1.625}\\
            w/o $\mathcal{L}_{penalty}$ & \textbf{32.61}  & \textbf{0.960}     &   6.625\\
            w/o $\mathcal{L}_{budg}$     & 32.02  & 0.957     &   5.875\\
            Ours                        & 32.21  & 0.957     &   3.625\\
            \bottomrule
    \end{tabular}
    
    \label{tab:advloss}
\end{table}

Table~\ref{tab:advloss} analyzes the components of the temporal budget loss. 
Removing $\mathcal{L}_{extra}$ drives the learned time step to 1.625, but the aggressive reduction causes a clear PSNR drop. Removing $\mathcal{L}_{penalty}$ gives the highest PSNR, 
while the average time step increases to 6.625, leaving limited room for energy reduction. 
Removing the complete $\mathcal{L}_{budg}$ also weakens time-step control and results in a larger temporal budget than PATA. The full objective reaches 32.21 dB PSNR with an average time step of 3.625, 
showing a balanced quality-efficiency trade-off under the $\beta=5\times10^{-7}$ setting.

\begin{table}[!t]
\caption{Comparison of different fixed target time-step settings.}
\label{tab:fixed_timestep}
\centering
\resizebox{\linewidth}{!}{
\begin{tabular}{lcccc}
\toprule
Method & Time Step $\downarrow$ & PSNR $\uparrow$ & SSIM $\uparrow$ & Energy (mJ) $\downarrow$ \\
\midrule
PATA & 3.625 & 32.21 & 0.957 & 492.25 \\
Fixed-$t$ & 3.625 & 32.11 & 0.956 & 491.70 \\
Fixed-$(t+1)$ & 4.625 & 32.29 & 0.958 & 623.01 \\
Fixed-$(t-1)$ & 2.625 & 31.69 & 0.954 & 398.37 \\
\bottomrule
\end{tabular}
}
\end{table}

Table~\ref{tab:fixed_timestep} compares PATA with fixed target-step settings in the second training stage. After obtaining the scene-wise target step $t_r^*$ learned by PATA, 
we fix the target step to $t_r^*$, $t_r^*+1$, and $t_r^*-1$ while keeping the other settings unchanged. Fixed-$t$ achieves nearly the same PSNR and SSIM as PATA, 
showing that the learned target step is a stable temporal budget. Increasing the target step gives only a small quality gain but raises the estimated energy, 
while decreasing it reduces energy at the cost of visible quality degradation. Since the fixed-$t$ setting uses the target step discovered by PATA, it serves as an oracle reference rather than a practical alternative.


\section{CONCLUSION}

In this work, we presented PATA, a two-stage adaptive time-step training framework for spike-based neural radiance fields. PATA learns a scene-wise temporal budget during training and applies a shared target time step to all sampling points within a scene, preserving parallel inference while adapting computation to scene complexity. The framework combines hybrid input dynamics, full-step pretraining, temporal-budget optimization, and smoothing regularization to reduce the inference time-step budget while maintaining rendering quality.

Experiments on INGP-NeRF and TensoRF show that PATA provides a controllable quality-efficiency trade-off across 
hash-grid and tensor-factorized neural rendering backbones. On INGP-NeRF, PATA reduces the estimated inference energy 
by up to 57.57\% on Synthetic-NeRF and 37.46\% on Mip-NeRF 360. On TensoRF, PATA is applied only to the color decoder 
while keeping the tensor-factorized representation and volume rendering pipeline unchanged, achieving up to 40.66\% estimated energy reduction on Synthetic-NeRF 
and 68.90\% on LLFF. These results show that scene-wise temporal-budget learning provides a practical route toward efficient spike-based neural rendering across different radiance-field representations.


\printcredits

\section*{Declaration of competing interest}
The authors declare that they have no known competing financial 
interests or personal relationships that could have appeared to influence 
the work reported in this paper.

\section*{Acknowledgments}

\bibliographystyle{cas-model2-names}
\bibliography{references}

\end{document}